\documentclass{article}

\usepackage{microtype}
\usepackage{graphicx}
\usepackage{subfigure}
\usepackage{booktabs} 

\usepackage{hyperref}

\usepackage[accepted]{style_files/ool2020}

\icmltitlerunning{Slot Contrastive Networks}

\begin{document}

\twocolumn[
\icmltitle{Slot Contrastive Networks: A Contrastive Approach for Representing Objects}




\begin{icmlauthorlist}
\icmlauthor{Evan Racah}{mila}
\icmlauthor{Sarath Chandar}{mila,poly}
\end{icmlauthorlist}

\icmlaffiliation{mila}{Mila, Montreal, QC, Canada}
\icmlaffiliation{poly}{Polytechnique Montreal, Montreal, QC}

\icmlcorrespondingauthor{Evan Racah}{ejracah@gmail.com}

\icmlkeywords{Machine Learning, ICML}

\vskip 0.3in
]



\printAffiliationsAndNotice{\icmlEqualContribution} 

\begin{abstract}
Unsupervised extraction of objects from low-level visual data is an important goal for further progress in machine learning. Existing approaches for representing objects without labels use structured generative models with static images. These methods focus a large amount of their capacity on reconstructing unimportant background pixels, missing low contrast or small objects. Conversely, we present a new method that avoids losses in pixel space and over-reliance on the limited signal a static image provides. Our approach takes advantage of objects' motion by learning a discriminative, time-contrastive loss in the space of slot representations, attempting to force each slot to not only capture entities that move, but capture distinct objects from the other slots. Moreover, we introduce a new quantitative evaluation metric to measure how ``diverse'' a set of slot vectors are, and use it to evaluate our model on 20 Atari games.
\end{abstract}

\section{Introduction}
Understanding a scene from low-level sensory data is an important aspect of human intelligence. One way humans are able to do this is by explicitly learning representations of objects in the scene. Effectively encoding objects is emerging as an important subfield in machine learning because it has the potential to lead to better representations, which accelerates the learning of tasks requiring understanding or interaction with objects and can potentially allow transfer to unseen tasks.  Furthermore, these structured object-like representations can be used as input to and are often a pre-requisite to structured downstream systems like graph-based relational techniques \cite{battaglia2016interaction}, casual modelling \cite{pearl2009causality}, physics-based simulators \cite{sanchez2020learning}, and reinforcement learning from low-dimensional state vectors.
There are many existing approaches for representing objects in computer vision with bounding boxes \citep{redmon2016you}; however, these approaches all require external supervision in the form of large numbers of human-labelled bounding box coordinates, which are expensive to obtain. To avoid the reliance on labels, many impressive unsupervised object representation approaches have been developed. However, most of these techniques involve generative models, which have two issues: wasted capacity on modelling spurious background pixels \cite{oord2018representation} and inability to capture small objects \cite{anand2019unsupervised}. As a promissing alternative to generative models for representation learning, discriminative, self-supervised techniques have emerged, which can be divided into two subcategories: pretext-task based techniques \cite{weng2019selfsup} and contrastive techniques \cite{anand2020contrastive,arora2019theoretical}. Indeed, many of the recent state of the art models for unsupervised pretraining on static image datasets involve these contrastive techniques \cite{bachman2019learning,henaff2019data,chen2020simple,he2019momentum}. While these results are impressive, they are designed to work with static images and not sequential visual datasets, like videos or transition tuples in reinforcement learning. This means that these methods miss out on the helpful signal that time provides, like the fact that interesting entities in a scene are often the ones that move or change with time. As a result, many self-supervised pretext approaches \cite{misra2016shuffle,aytar2018playing} and contrastive approaches \cite{hyvarinen2017nonlinear,oord2018representation,anand2019unsupervised} have begun to harness time in their self-supervised signal. However, these approaches are often unstructured in the sense that they model the scene with just one global vector instead of a set of representation vectors of separate entities. As a result, we aim to learn structured, object-centric slot representations harnessing time and using a self-supervised time-contrastive signal similar to \cite{anand2019unsupervised,hyvarinen2017nonlinear} to learn each object's representation, but also a ``slot contrastive'' signal as an attempt to force each slot to capture a unique object compared to the other slots.



\section{Slot Contrastive Networks}
\subsection{Architecture}
\begin{figure}[ht]
\vskip 0.2in
\begin{center}
\centerline{\includegraphics[width=\columnwidth]{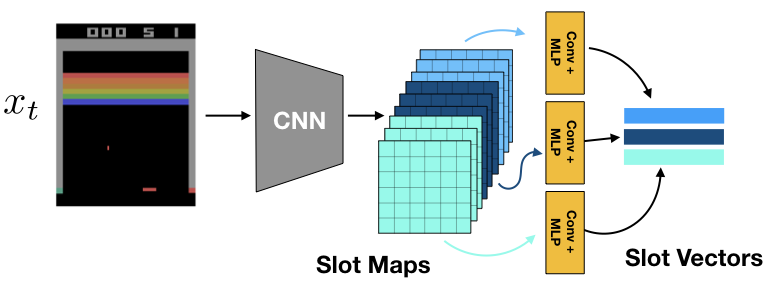}}
\caption{The architecture of Slot Contrastive Networks (SCN). A single frame is encoded with a CNN and then the feature maps are divided into groups called slot maps and separately passed through a shared-weight convolutional layer plus MLP to produce a set of  slot vectors.}
\label{arch-scn}
\end{center}
\vskip -0.2in
\end{figure}
The architecture of slot contrastive networks, as shown in Figure \ref{arch-scn}, is structured as a standard convolutional neural network, but where instead of the network encoding a single frame into one vector representation, it encodes the frame, $x_t$ into $K$ slot vectors, $s_{t}^i; i\in{1,2,3,...,k}$. It does this by splitting up the feature maps from the CNN into $K$ sets of feature maps, which we call ``slot maps'', each separately encoded into a different slot vector by a small sub-network (convolutional layer followed by MLP) with shared weights.  
\subsection{Losses}
\begin{figure}[ht]
\vskip 0.2in
\begin{center}
\centerline{\includegraphics[width=\columnwidth]{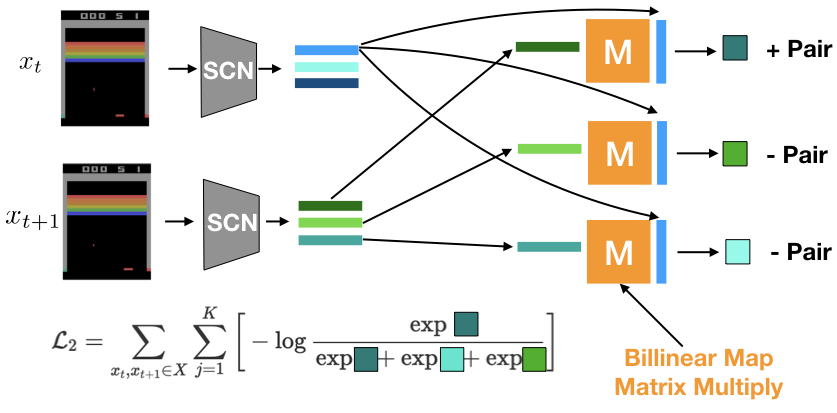}}
\caption{The second loss term, encouraging slot diversity, shown for a single slot.}
\label{fig-loss2}
\end{center}
\vskip -0.2in
\end{figure}

The losses of SCN are computed by separately encoding frames from consecutive time steps into slot vectors and then computing relationships between the slot vectors. The loss has two terms that attempt to enforce two constraints on the slot representation: slot saliency and slot diversity.
\paragraph{Loss Term 1 ($\mathcal{L}_1$): Encouraging Slot Saliency}
With slot saliency we want each slot vector to capture an important part of the scene, namely an object. To enforce that objective, we take advantage of our main assumption that objects and other important parts of a scene change in time, while spurious parts like backgrounds do not. Thus, we formulate a loss that tries to ensure that the slot representations capture ``time dependent'' features (i.e. capture things that move). This is the intuition that motivates time-contrastive losses \cite{hyvarinen2017nonlinear,anand2019unsupervised,sermanet2018time}; learning state representations that make it easy to predict the temporal distance between states, will potentially ensure that these representations capture time dependent features. We adapt this type of loss to slot-structured representations by designing an InfoNCE loss \cite{oord2018representation} to contrast similar or positive pairs (the same slot at two consecutive time steps) with dissimilar or negative pairs (the same slot at random, likely nonconsecutive, time steps). The loss shown in Equation \ref{eq-loss1} ends up looking similar to a standard softmax multiclass classification loss, so we can describe it as classifying a positive pair among many negative pairs.
\begin{equation}
\label{eq-loss1}
    \mathcal{L}_1 = \sum_{x_t,x_{t+1} \in X}
           \sum_{j=1}^{K} 
                    \bigg[ -\log \frac{ \exp f_{jj}(x_t, x_{t+1})}
                    { \sum_{x_{t'}\in X_{+1}} \exp f_{jj}(x_t,x_{t'}) } \bigg] 
    \end{equation}
where $X = \{ (x_t, x_{t+1})_i \}_{i=1}^N$ is a minibatch of consecutive pairs of frames that are randomly sampled from collected episodes and $X_{+1} = X[:, 1]$ is the second element of the pair from the set of pairs in the minibatch. In addition $f_{ij}(X_1,X_2)$ is a function of the $i$-th slot from frame $X_1$ and the $j$-th slot from frame $X_2$. In our case it is a bilinear map: $f_{ij}(X_1,X_2) = \phi_i(X_1)^T W \phi_j(X_2)$, where $\phi_j(X_2)$ is the function that extracts the $j$-th slot from frame $X_2$ and $W$ is a matrix of size $|s_t^i| \times |s_t^i|$, where $|s_t^i|$ is the length of a slot vector.
\paragraph{Loss Term 2 ($\mathcal{L}_2$): Encouraging Slot Diversity}
To encourage diversity between slots, we try to incentivize each slot representation to be different from the others. We achieve this by implementing a ``slot contrastive'' loss, where we train a classifier to predict whether a pair of slot representations consists of the same slot at consecutive time steps or if the pair consists of representations from two different slots. We again implement this as a contrastive loss using InfoNCE as shown in Equation \ref{eq-loss2} and Figure \ref{fig-loss2}
\begin{equation}
\label{eq-loss2}
\mathcal{L}_2 = \sum_{x_t,x_{t+1} \in X}
        \sum_{j=1}^{K} \bigg[ -\log \frac{ \exp f_{jj}(x_t,x_{t+1})} {\sum_{i=1}^{K} \exp f_{ji}(x_t,x_{t+1}) } \bigg] 
\end{equation}

\section{Related Work}
\label{rel-work}
There have been many previous approaches for unsupervised learning of object-centric representations. Most previous works have involved various latent generative models. The approaches differ in the structure and assumptions they impose on their models. The first type is spatial attention models which attend different locations in the scene to extract objects \citep{kosiorek2018sequential,eslami2016attend,crawford2019exploiting,lin2020space,jiang2019scalable} and the second is scene-mixture models, where the scene is modelled as a Gaussian mixture model of scene components \citep{nash2017multi,greff2016tagger,greff2017neural,greff2019multi,burgess2019monet}. The third major form of object-centric models are keypoint models \citep{zhang2018unsupervised,jakab2018unsupervised}, which extract keypoints (the spatial coordinates of entities) by fitting 2D Gaussians to the feature maps of an encoder-decoder model. In addition, some works have begun to use video-like datasets, so that objects can be extracted by harnessing their movement. For example, several works have jointly extracted objects and computed their interactions with each other to predict their future state for: scene mixture models \cite{van2018relational,engelcke2019genesis}, spatial attention models \cite{kossen2019structured}, and keypoint models \citep{kulkarni2019unsupervised,minderer2019unsupervised}. All of these models are trained to reconstruct the input scene in pixel space. In contrast, a few works have begun using discriminative models for learning objects including \citep{ehrhardt2018unsupervised}, which uses a temporal self-supervised pretext task to learn objects and constrastive structured world models (CSWM) \cite{kipf2019contrastive}, which predicts future object representations with a contrastive training loss.
\paragraph{Similarities and differences to CSWM}
While both our model and CSWM use a contrastive loss in the space of the slot representations, and use time to provide a notion of similarity, there are a few key differences. For instance, for each slot at a given time step, CSWM predicts that slot's representation at the next time step using a graph neural network, while our model can be thought of as using a linear layer. Their distance function between pairs is Euclidean distance, while ours is a dot product. CSWM uses a hinge-based formulation to maximize the positive pair distance and minimize the negative pair distance, while we use InfoNCE \cite{oord2018representation}. Lastly, while CSWM's loss is an intra-slot loss, which bears many similarities to our first loss term, we add a novel inter-slot loss term (for encouraging slot diversity) which has no analog in CSWM.
 \section{Evaluating Slot Representations}
 Traditionally, slot representations have been evaluated by inspecting qualitative reconstructions \cite{greff2019multi} or measuring how similar temporally close slots are in representation space \cite{kipf2019contrastive}. However, there have not been many quantitative measures of slot representations grounded in how well they capture the true state of objects in the scene. Borrowing from the self-supervision and disentangling literature, we propose several evaluation metrics to measure how accurately slots capture objects in the scene and how disentangled each slot is from the others. We use three measures from the disentangling community, which we adapt to slot representations: slot accuracy (sometimes called explicitness), slot modularity, and slot compactness.
\paragraph{Slot Accuracy}
For slot accuracy, we use linear probing, a technique commonly used in self-supervised learning \cite{anand2019unsupervised,hjelm2018learning,chen2020simple} and disentangling \cite{locatello2018challenging}.
 We concatenate all the slots into one vector and then input it to multiple linear regressors each trained to regress a particular x or y coordinate of a particular object in the scene. Accuracy values are the $R^2$ score of each linear regressor. Negative values for $R^2$ are possible, which happens when the total squared error of the linear regressor is more than the variance of the ground truth coordinate values.
\paragraph{Slot Compactness and Slot Modularity}
Inspired by \citep{eastwood2018framework}, we compute a variant of DCI completeness \cite{locatello2018challenging}, which we call \textbf{slot compactness}. To compute slot compactness, we first take the weights of the linear regressor probes used to compute slot accuracy, then we take their absolute value and normalize them to create a feature importance matrix denoting how ``important'' each element of each slot vector is to regressing each object's coordinate. We then average the feature importances across each slot to get a slot importance matrix, which has shape $P \times K$, where $K$ is the number of slots and $P$ is the number of objects. The element at index $(i, j)$ denotes how important the $i$-th slot is for encoding the $j$-th object.  We then treat each row as a probability distribution and compute the average of one minus the entropy of each row of the matrix to get the slot compactness. This gives a score between $0$ and $1$, where the higher score the fewer slots contribute to encoding an object. \textbf{Slot modularity}, which is inspired by DCI disentangling \citep{eastwood2018framework,locatello2018challenging} is computed by calculating one minus the entropy of each column of the slot importance matrix (after normalizing the columns). This gives a score between $0$ and $1$, where the higher score the fewer objects are encoded by a slot.
\section{Experiments}
For our experiments we train our slot contrastive networks using full-sized $210 \times 160$ RGB frames from 20 Atari games. For evaluation, we use labels from the AtariARI dataset \cite{anand2019unsupervised}, restricting ourselves to labels that correspond to the x or y coordinates of objects. We set $K$ to be equal to the true number of objects in the game, which is a common practice used in \cite{kipf2019contrastive,kulkarni2019unsupervised}. Following \cite{anand2019unsupervised}, we train our model with 100,000 frames acquired with a random agent on the Atari games; an additional 50,000 frames are used for training and testing the evaluation probes. We compare to several baselines: a randomly initialized model with no training, CSWM \citep{kipf2019contrastive} and a fully supervised model, where each slot in the model is trained to regress the true position of one of the objects in the scene. The architectural details of all models are similar to \citep{anand2019unsupervised}. We use 20 of the 22 games in AtariARI; we skipped Hero because it only contains one object and Qbert because of poor regression performance even for the supervised model (negative $R^2$ values).
\begin{table}[t]
\caption{\textbf{Slot Compactness and Slot Modularity Scores}: Averaged across 20 Atari games for slot compactness and 19 games for slot modularity (Bowling only has one object). Higher is better}
\label{slot-div-table}
\vskip 0.15in
\begin{center}
\begin{small}
\begin{sc}
\scalebox{0.75}{
\begin{tabular}{lrrrr}
\toprule
{} &  random-cnn &    scn &   cswm &  supervised \\
\midrule
Slot Modularity &  0.003  &  0.004  &  \textbf{0.041}   &  0.198\\
Slot Compactness &  0.007  &  0.014  &  \textbf{0.304}   &  0.266\\
\bottomrule
\end{tabular}}
\end{sc}
\end{small}
\end{center}
\vskip -0.1in
\end{table}
\begin{table}[t]
\caption{\textbf{Ablation Results} across all 20 games for the ablation of SCN where the second loss term is removed. Higher is better}
\label{slot-acc-abl}
\begin{center}
\begin{small}
\begin{sc}
\begin{tabular}{lrr}
\toprule
{} &  scn\_loss1only &    scn \\
\midrule
Slot Accuracy           &  0.40  &  \textbf{0.45} \\
Slot Modularity   &  0.0041  &  \textbf{0.0045} \\
Slot Compactness  &  \textbf{0.0170}   &  0.0137\\
\bottomrule
\end{tabular}
\end{sc}
\end{small}
\end{center}
\vskip -0.1in
\end{table}
\begin{table}[t]
\caption{\textbf{Slot Accuracy} average $R^2$ score for linear regressors regressing object coordinates trained on all slots concatenated for all 20 games. }
\label{slot-acc}
\vskip 0.15in
\begin{center}
\begin{small}
\begin{sc}
\scalebox{0.70}{
\begin{tabular}{lrrrr}
\toprule
{} &  random-cnn &     scn &    cswm &  supervised \\
\midrule
Asteroids          &  \textbf{0.14}   &  0.11  &  0.11  &  0.39\\
Berzerk            &  0.30  &  0.35  &  \textbf{0.38}   &  0.66\\
Bowling            &  0.39  &  0.83  &  \textbf{0.96}   &  1.00\\
Boxing             &  \textbf{0.71}   &  0.68  &  0.35  &  1.00\\
Breakout           &  0.21  &  0.55  &  \textbf{0.70}   &  0.75\\
Demonattack        &  0.10  &  \textbf{0.22}   &  0.14  &  0.72\\
Freeway            &  0.58  &  \textbf{0.83}   &  \textbf{0.84}   &  0.98\\
Frostbite          &  \textbf{0.71}   &  0.48  &  0.69  &  0.94\\
Montezumarevenge   &  0.70  &  0.79  &  \textbf{0.92}   &  0.99\\
Mspacman           &  \textbf{0.13}   &  0.11  &  -0.04  &  0.82\\
Pitfall            &  \textbf{0.50}   &  0.27  &  0.30  &  0.83\\
Pong               &  0.47  &  0.73  &  \textbf{0.82}   &  0.93\\
Privateeye         &  \textbf{0.74}   &  0.57  &  0.39  &  0.99\\
Riverraid          &  0.11  &  0.38  &  \textbf{0.59}   &  0.94\\
Seaquest           &  0.36  &  0.39  &  \textbf{0.54}   &  0.83\\
Spaceinvaders      &  \textbf{0.41}   &  \textbf{0.42}   &  0.21  &  0.92\\
Tennis             &  0.43  &  0.50  &  \textbf{0.66}   &  0.90\\
Venture            &  0.12  &  0.12  &  \textbf{0.27}   &  0.50\\
Videopinball       &  \textbf{0.48}   &  0.45  &  0.01  &  0.99\\
Yarsrevenge        &  \textbf{0.17}   &  0.11  &  0.11  &  0.81\\
Overall            &  0.39  &  \textbf{0.45}   &  \textbf{0.45}   &  0.84\\
\bottomrule
\end{tabular}}
\end{sc}
\end{small}
\end{center}
\vskip -0.1in
\end{table}

\section{Discussion}
Interestingly enough, CSWM and SCN perform similarly in average slot accuracy across all games. CSWM particularly excels at games with a few objects that interact frequently, like Pong and Breakout, which both have a ball bouncing off of a paddle. CSWM also seems perform well at games with very predictable, repeatable motion, like the cars in Freeway and the fish in Seaquest. This is likely because CSWM is trained to learn features that minimize its prediction error. CSWM struggles and SCN performs better in games where the motion is not as regular and predictable, like Boxing and VideoPinball. This may be because SCN trained to find any objects that move regardless of if they are easily predictable. One strange result is CSWM's negative $R^2$ score for Ms. Pacman. This could be because the diversity of frames one obtains with a random policy on these games is small; for example, the agent in Ms. Pacman will basically stay in one place in expectation and as a result the ghosts, who follow the agent, will not move around much. It is interesting to note from Table \ref{slot-div-table} that SCN's slots are not as modular or compact as CSWM. This evidence suggests that potentially the slot diversity loss term in SCN has little actual effect on slot diversity. This is demonstrated more explicitly in table \ref{slot-acc-abl}, which shows that removing the slot diversity term of SCN results in almost no noticeable change in modularity and actually a small improvement in compactness. The lack of slot diversity diminishment paired with a decrease in slot accuracy when removing the slot diversity term in SCN suggests it provides a slight regularization benefit, but little else. Perhaps, each slot is focusing on the same object, but different parts of it. Without enforcing any true spatial disentangling between slots, it may be hard to truly coax the slots capturing different objects.
\subsection{Future Work}
These results suggest that perhaps a simple future direction could be a more careful tuning of a coefficient of the second term of the loss is needed. However, a potentially more elegant solution is to try to force slot diversity through architectural inductive biases instead of loss objectives. For instance, a hard or soft spatial attention or routing module for each slot would be an interesting future direction to pair with the time constrastive objective in slot space. Lastly, exploring ways to enhance this model with the ability to dynamically determine the number of slots. Some of these future directions could be thought of as pairing spatial attention models, like \cite{crawford2019spatially,jiang2019scalable,lin2020space}, with a temporal contrastive loss instead of a reconstruction loss.



\pagebreak
\bibliography{biblio}

\begin{thebibliography}{40}
\providecommand{\natexlab}[1]{#1}
\providecommand{\url}[1]{\texttt{#1}}
\expandafter\ifx\csname urlstyle\endcsname\relax
  \providecommand{\doi}[1]{doi: #1}\else
  \providecommand{\doi}{doi: \begingroup \urlstyle{rm}\Url}\fi

\bibitem[Anand(2020)]{anand2020contrastive}
Anand, A.
\newblock Contrastive self-supervised learning, 2020.
\newblock
  \url{https://ankeshanand.com/blog/2020/01/26/contrative-self-supervised-learning.html}.

\bibitem[Anand et~al.(2019)Anand, Racah, Ozair, Bengio, C{\^o}t{\'e}, and
  Hjelm]{anand2019unsupervised}
Anand, A., Racah, E., Ozair, S., Bengio, Y., C{\^o}t{\'e}, M.-A., and Hjelm,
  R.~D.
\newblock Unsupervised state representation learning in atari.
\newblock In \emph{Advances in Neural Information Processing Systems}, pp.\
  8766--8779, 2019.

\bibitem[Arora et~al.(2019)Arora, Khandeparkar, Khodak, Plevrakis, and
  Saunshi]{arora2019theoretical}
Arora, S., Khandeparkar, H., Khodak, M., Plevrakis, O., and Saunshi, N.
\newblock A theoretical analysis of contrastive unsupervised representation
  learning.
\newblock \emph{arXiv preprint arXiv:1902.09229}, 2019.

\bibitem[Aytar et~al.(2018)Aytar, Pfaff, Budden, Paine, Wang, and
  de~Freitas]{aytar2018playing}
Aytar, Y., Pfaff, T., Budden, D., Paine, T.~L., Wang, Z., and de~Freitas, N.
\newblock Playing hard exploration games by watching youtube.
\newblock \emph{arXiv preprint arXiv:1805.11592}, 2018.

\bibitem[Bachman et~al.(2019)Bachman, Hjelm, and
  Buchwalter]{bachman2019learning}
Bachman, P., Hjelm, R.~D., and Buchwalter, W.
\newblock Learning representations by maximizing mutual information across
  views.
\newblock In \emph{Advances in Neural Information Processing Systems}, pp.\
  15509--15519, 2019.

\bibitem[Battaglia et~al.(2016)Battaglia, Pascanu, Lai, Rezende,
  et~al.]{battaglia2016interaction}
Battaglia, P., Pascanu, R., Lai, M., Rezende, D.~J., et~al.
\newblock Interaction networks for learning about objects, relations and
  physics.
\newblock In \emph{Advances in neural information processing systems}, pp.\
  4502--4510, 2016.

\bibitem[Burgess et~al.(2019)Burgess, Matthey, Watters, Kabra, Higgins,
  Botvinick, and Lerchner]{burgess2019monet}
Burgess, C.~P., Matthey, L., Watters, N., Kabra, R., Higgins, I., Botvinick,
  M., and Lerchner, A.
\newblock Monet: Unsupervised scene decomposition and representation.
\newblock \emph{arXiv preprint arXiv:1901.11390}, 2019.

\bibitem[Chen et~al.(2020)Chen, Kornblith, Norouzi, and Hinton]{chen2020simple}
Chen, T., Kornblith, S., Norouzi, M., and Hinton, G.
\newblock A simple framework for contrastive learning of visual
  representations.
\newblock \emph{arXiv preprint arXiv:2002.05709}, 2020.

\bibitem[Crawford \& Pineau(2019{\natexlab{a}})Crawford and
  Pineau]{crawford2019exploiting}
Crawford, E. and Pineau, J.
\newblock Exploiting spatial invariance for scalable unsupervised object
  tracking.
\newblock \emph{arXiv preprint arXiv:1911.09033}, 2019{\natexlab{a}}.

\bibitem[Crawford \& Pineau(2019{\natexlab{b}})Crawford and
  Pineau]{crawford2019spatially}
Crawford, E. and Pineau, J.
\newblock Spatially invariant unsupervised object detection with convolutional
  neural networks.
\newblock In \emph{Proceedings of the AAAI Conference on Artificial
  Intelligence}, volume~33, pp.\  3412--3420, 2019{\natexlab{b}}.

\bibitem[Eastwood \& Williams(2018)Eastwood and
  Williams]{eastwood2018framework}
Eastwood, C. and Williams, C.~K.
\newblock A framework for the quantitative evaluation of disentangled
  representations.
\newblock 2018.

\bibitem[Ehrhardt et~al.(2018)Ehrhardt, Monszpart, Mitra, and
  Vedaldi]{ehrhardt2018unsupervised}
Ehrhardt, S., Monszpart, A., Mitra, N., and Vedaldi, A.
\newblock Unsupervised intuitive physics from visual observations.
\newblock In \emph{Asian Conference on Computer Vision}, pp.\  700--716.
  Springer, 2018.

\bibitem[Engelcke et~al.(2019)Engelcke, Kosiorek, Jones, and
  Posner]{engelcke2019genesis}
Engelcke, M., Kosiorek, A.~R., Jones, O.~P., and Posner, I.
\newblock Genesis: Generative scene inference and sampling with object-centric
  latent representations.
\newblock \emph{arXiv preprint arXiv:1907.13052}, 2019.

\bibitem[Eslami et~al.(2016)Eslami, Heess, Weber, Tassa, Szepesvari, Hinton,
  et~al.]{eslami2016attend}
Eslami, S.~A., Heess, N., Weber, T., Tassa, Y., Szepesvari, D., Hinton, G.~E.,
  et~al.
\newblock Attend, infer, repeat: Fast scene understanding with generative
  models.
\newblock In \emph{Advances in Neural Information Processing Systems}, pp.\
  3225--3233, 2016.

\bibitem[Greff et~al.(2016)Greff, Rasmus, Berglund, Hao, Valpola, and
  Schmidhuber]{greff2016tagger}
Greff, K., Rasmus, A., Berglund, M., Hao, T., Valpola, H., and Schmidhuber, J.
\newblock Tagger: Deep unsupervised perceptual grouping.
\newblock In \emph{Advances in Neural Information Processing Systems}, pp.\
  4484--4492, 2016.

\bibitem[Greff et~al.(2017)Greff, Van~Steenkiste, and
  Schmidhuber]{greff2017neural}
Greff, K., Van~Steenkiste, S., and Schmidhuber, J.
\newblock Neural expectation maximization.
\newblock In \emph{Advances in Neural Information Processing Systems}, pp.\
  6691--6701, 2017.

\bibitem[Greff et~al.(2019)Greff, Kaufmann, Kabra, Watters, Burgess, Zoran,
  Matthey, Botvinick, and Lerchner]{greff2019multi}
Greff, K., Kaufmann, R.~L., Kabra, R., Watters, N., Burgess, C., Zoran, D.,
  Matthey, L., Botvinick, M., and Lerchner, A.
\newblock Multi-object representation learning with iterative variational
  inference.
\newblock \emph{arXiv preprint arXiv:1903.00450}, 2019.

\bibitem[He et~al.(2019)He, Fan, Wu, Xie, and Girshick]{he2019momentum}
He, K., Fan, H., Wu, Y., Xie, S., and Girshick, R.
\newblock Momentum contrast for unsupervised visual representation learning.
\newblock \emph{arXiv preprint arXiv:1911.05722}, 2019.

\bibitem[H{\'e}naff et~al.(2019)H{\'e}naff, Srinivas, De~Fauw, Razavi, Doersch,
  Eslami, and Oord]{henaff2019data}
H{\'e}naff, O.~J., Srinivas, A., De~Fauw, J., Razavi, A., Doersch, C., Eslami,
  S., and Oord, A. v.~d.
\newblock Data-efficient image recognition with contrastive predictive coding.
\newblock \emph{arXiv preprint arXiv:1905.09272}, 2019.

\bibitem[Hjelm et~al.(2018)Hjelm, Fedorov, Lavoie-Marchildon, Grewal, Bachman,
  Trischler, and Bengio]{hjelm2018learning}
Hjelm, R.~D., Fedorov, A., Lavoie-Marchildon, S., Grewal, K., Bachman, P.,
  Trischler, A., and Bengio, Y.
\newblock Learning deep representations by mutual information estimation and
  maximization.
\newblock \emph{arXiv preprint arXiv:1808.06670}, 2018.

\bibitem[Hyvarinen \& Morioka(2017)Hyvarinen and
  Morioka]{hyvarinen2017nonlinear}
Hyvarinen, A. and Morioka, H.
\newblock Nonlinear ica of temporally dependent stationary sources.
\newblock Proceedings of Machine Learning Research, 2017.

\bibitem[Jakab et~al.(2018)Jakab, Gupta, Bilen, and
  Vedaldi]{jakab2018unsupervised}
Jakab, T., Gupta, A., Bilen, H., and Vedaldi, A.
\newblock Unsupervised learning of object landmarks through conditional image
  generation.
\newblock In \emph{Advances in Neural Information Processing Systems}, pp.\
  4016--4027, 2018.

\bibitem[Jiang et~al.(2019)Jiang, Janghorbani, de~Melo, and
  Ahn]{jiang2019scalable}
Jiang, J., Janghorbani, S., de~Melo, G., and Ahn, S.
\newblock Scalable object-oriented sequential generative models.
\newblock \emph{arXiv preprint arXiv:1910.02384}, 2019.

\bibitem[Kipf et~al.(2019)Kipf, van~der Pol, and Welling]{kipf2019contrastive}
Kipf, T., van~der Pol, E., and Welling, M.
\newblock Contrastive learning of structured world models.
\newblock \emph{arXiv preprint arXiv:1911.12247}, 2019.

\bibitem[Kosiorek et~al.(2018)Kosiorek, Kim, Teh, and
  Posner]{kosiorek2018sequential}
Kosiorek, A., Kim, H., Teh, Y.~W., and Posner, I.
\newblock Sequential attend, infer, repeat: Generative modelling of moving
  objects.
\newblock In \emph{Advances in Neural Information Processing Systems}, pp.\
  8606--8616, 2018.

\bibitem[Kossen et~al.(2019)Kossen, Stelzner, Hussing, Voelcker, and
  Kersting]{kossen2019structured}
Kossen, J., Stelzner, K., Hussing, M., Voelcker, C., and Kersting, K.
\newblock Structured object-aware physics prediction for video modeling and
  planning.
\newblock \emph{arXiv preprint arXiv:1910.02425}, 2019.

\bibitem[Kulkarni et~al.(2019)Kulkarni, Gupta, Ionescu, Borgeaud, Reynolds,
  Zisserman, and Mnih]{kulkarni2019unsupervised}
Kulkarni, T.~D., Gupta, A., Ionescu, C., Borgeaud, S., Reynolds, M., Zisserman,
  A., and Mnih, V.
\newblock Unsupervised learning of object keypoints for perception and control.
\newblock In \emph{Advances in Neural Information Processing Systems}, pp.\
  10723--10733, 2019.

\bibitem[Lin et~al.(2020)Lin, Wu, Peri, Sun, Singh, Deng, Jiang, and
  Ahn]{lin2020space}
Lin, Z., Wu, Y.-F., Peri, S.~V., Sun, W., Singh, G., Deng, F., Jiang, J., and
  Ahn, S.
\newblock Space: Unsupervised object-oriented scene representation via spatial
  attention and decomposition.
\newblock \emph{arXiv preprint arXiv:2001.02407}, 2020.

\bibitem[Locatello et~al.(2018)Locatello, Bauer, Lucic, R{\"a}tsch, Gelly,
  Sch{\"o}lkopf, and Bachem]{locatello2018challenging}
Locatello, F., Bauer, S., Lucic, M., R{\"a}tsch, G., Gelly, S., Sch{\"o}lkopf,
  B., and Bachem, O.
\newblock Challenging common assumptions in the unsupervised learning of
  disentangled representations.
\newblock \emph{arXiv preprint arXiv:1811.12359}, 2018.

\bibitem[Minderer et~al.(2019)Minderer, Sun, Villegas, Cole, Murphy, and
  Lee]{minderer2019unsupervised}
Minderer, M., Sun, C., Villegas, R., Cole, F., Murphy, K.~P., and Lee, H.
\newblock Unsupervised learning of object structure and dynamics from videos.
\newblock In \emph{Advances in Neural Information Processing Systems}, pp.\
  92--102, 2019.

\bibitem[Misra et~al.(2016)Misra, Zitnick, and Hebert]{misra2016shuffle}
Misra, I., Zitnick, C.~L., and Hebert, M.
\newblock Shuffle and learn: unsupervised learning using temporal order
  verification.
\newblock In \emph{European Conference on Computer Vision}, pp.\  527--544.
  Springer, 2016.

\bibitem[Nash et~al.(2017)Nash, Eslami, Burgess, Higgins, Zoran, Weber, and
  Battaglia]{nash2017multi}
Nash, C., Eslami, S.~A., Burgess, C., Higgins, I., Zoran, D., Weber, T., and
  Battaglia, P.
\newblock The multi-entity variational autoencoder.
\newblock In \emph{NIPS Workshops}, 2017.

\bibitem[Oord et~al.(2018)Oord, Li, and Vinyals]{oord2018representation}
Oord, A. v.~d., Li, Y., and Vinyals, O.
\newblock Representation learning with contrastive predictive coding.
\newblock \emph{arXiv preprint arXiv:1807.03748}, 2018.

\bibitem[Pearl(2009)]{pearl2009causality}
Pearl, J.
\newblock \emph{Causality}.
\newblock Cambridge university press, 2009.

\bibitem[Redmon et~al.(2016)Redmon, Divvala, Girshick, and
  Farhadi]{redmon2016you}
Redmon, J., Divvala, S., Girshick, R., and Farhadi, A.
\newblock You only look once: Unified, real-time object detection.
\newblock In \emph{Proceedings of the IEEE conference on computer vision and
  pattern recognition}, pp.\  779--788, 2016.

\bibitem[Sanchez-Gonzalez et~al.(2020)Sanchez-Gonzalez, Godwin, Pfaff, Ying,
  Leskovec, and Battaglia]{sanchez2020learning}
Sanchez-Gonzalez, A., Godwin, J., Pfaff, T., Ying, R., Leskovec, J., and
  Battaglia, P.~W.
\newblock Learning to simulate complex physics with graph networks.
\newblock \emph{arXiv preprint arXiv:2002.09405}, 2020.

\bibitem[Sermanet et~al.(2018)Sermanet, Lynch, Chebotar, Hsu, Jang, Schaal,
  Levine, and Brain]{sermanet2018time}
Sermanet, P., Lynch, C., Chebotar, Y., Hsu, J., Jang, E., Schaal, S., Levine,
  S., and Brain, G.
\newblock Time-contrastive networks: Self-supervised learning from video.
\newblock In \emph{2018 IEEE International Conference on Robotics and
  Automation (ICRA)}, pp.\  1134--1141. IEEE, 2018.

\bibitem[Van~Steenkiste et~al.(2018)Van~Steenkiste, Chang, Greff, and
  Schmidhuber]{van2018relational}
Van~Steenkiste, S., Chang, M., Greff, K., and Schmidhuber, J.
\newblock Relational neural expectation maximization: Unsupervised discovery of
  objects and their interactions.
\newblock \emph{arXiv preprint arXiv:1802.10353}, 2018.

\bibitem[Weng(2019)]{weng2019selfsup}
Weng, L.
\newblock Self-supervised representation learning.
\newblock \emph{lilianweng.github.io/lil-log}, 2019.
\newblock URL
  \url{https://lilianweng.github.io/lil-log/2019/11/10/self-supervised-learning.html}.

\bibitem[Zhang et~al.(2018)Zhang, Guo, Jin, Luo, He, and
  Lee]{zhang2018unsupervised}
Zhang, Y., Guo, Y., Jin, Y., Luo, Y., He, Z., and Lee, H.
\newblock Unsupervised discovery of object landmarks as structural
  representations.
\newblock In \emph{Proceedings of the IEEE Conference on Computer Vision and
  Pattern Recognition}, pp.\  2694--2703, 2018.

\end{thebibliography}
\bibliographystyle{style_files/ool2020}


\end{document}